\documentclass{article}

\usepackage{PRIMEarxiv}
\usepackage[utf8]{inputenc} % allow utf-8 input
\usepackage[T1]{fontenc}    % use 8-bit T1 fonts
\usepackage{hyperref}      % hyperlinks
\usepackage{url}            % simple URL typesetting
\usepackage{booktabs}       % professional-quality tables
\usepackage{amsfonts}       % blackboard math symbols
\usepackage{nicefrac}       % compact symbols for 1/2, etc.
\usepackage{microtype}      % microtypography
\usepackage{lipsum}
\usepackage{fancyhdr}       % header
\usepackage{graphicx}       % graphics
\usepackage{subcaption}
\usepackage{tikz,lmodern}
\usepackage{natbib}
\usepackage[most]{tcolorbox}
\usepackage{algorithm}
\usepackage[noend]{algpseudocode}
\graphicspath{{media/}}     % organize your images and other figures under media/ folder

\title{Evaluating Stability of Unreflective Alignment}

\author{
  James Lucassen \\
   \And
  Mark Henry \\
   \And
  Philippa Wright
   \And
  Owen Yeung
}

\begin{document}
\maketitle

\begin{abstract}
Many theoretical obstacles to AI alignment are consequences of reflective stability - the problem of designing alignment mechanisms that the AI would not disable if given the option. However, problems stemming from reflective stability are not obviously present in current LLMs, leading to disagreement over whether they will need to be solved to enable safe delegation of cognitive labor. In this paper, we propose Counterfactual Priority Change (CPC) destabilization as a mechanism by which reflective stability problems may arise in future LLMs. We describe two risk factors for CPC-destabilization: 1) CPC-based stepping back and 2) preference instability. We develop preliminary evaluations for each of these risk factors, and apply them to frontier LLMs. Our findings indicate that in current LLMs, increased scale and capability are associated with increases in both CPC-based stepping back and preference instability, suggesting that CPC-destabilization may cause reflective stability problems in future LLMs.
\end{abstract}

\begin{itemize}
    \item \url{https://github.com/jlucassen/CPC_stepping_back}
    \item \url{https://github.com/jlucassen/value_instability}
\end{itemize}

\section{Introduction}
\subsection{Motivation}
The goal of this work is to help inform prioritization of AI alignment research. To that end, we investigate what sorts of alignment problems are likely to affect future generations of LLM-based AI. Specifically, we focus on an open question whose answer has significant implications for problem prioritization - “will alignment need to be reflectively stable before useful cognitive labor can be safely delegated to AI?”

A property of an AI system is reflectively stable if, when given the choice to modify itself, the AI preserves that property\citep{1}. The alignment of an AI is reflective stable if, given the choice to modify itself, the AI remains aligned to the same goals. All else equal, reflectively stable alignment is preferable, as it rules out failure modes where alignment properties “uninstall themselves”. However, it seems that reflectively stable alignment may be difficult to achieve. Early research in the context of expected utility maximization failed to identify utility functions that are reflectively stable while also being indifferent to shutdown and pursuing some external utility term\citep{2}. More recent work has identified a potential approach using preferential gaps, but this approach requires extremely robust generalization of preference structure, and it remains unclear how tractable it is\citep{3}. On the other hand, alignment techniques such as RLHF\citep{4} which do not attempt reflective stability have recently demonstrated significant progress in managing the behavior of present frontier LLMs.

This raises a very natural question - just how necessary is reflectively stable alignment? Can we do without it? A prominent approach\citep{5} to AI safety aims to solve only the problems needed to safely delegate the remaining alignment research to AI. Current LLMs have made rapid progress on a variety of cognitive tasks\citep{6}, and do not yet seem to develop alignment failures due to reflective instability. Perhaps by the time we reach the capability threshold needed to delegate alignment research to AI, alignment failures due to reflective instability will still not have materialized.

Events may move very fast when AI capabilities become advanced enough to automate entire fields of research such as AI alignment. It would be best to know in advance whether or not such AIs will need to be reflectively stable in order to remain aligned, rather than waiting to find out. However, the theoretical debate on this topic is at something of an impasse. We hope to make progress by investigating this question empirically. To operationalize the question, we propose a concrete pathway that may cause alignment failures due to reflective instability in LLMs before they are able to automate AI alignment research, and investigate risk factors for this particular threat.

\subsection{CPC-Destabilization}

“Long-horizon” tasks are those which require an ongoing, iterative process or extended thinking to solve. Some extremely simple examples of long-horizon tasks might include Tic Tac Toe and Connect Four\citep{7}, in contrast to short-horizon tasks like answering a multiple-choice trivia question or squaring a number. Implementing a single function from a verbal description might be “short-horizon”, but implementing an entire codebase with many such functions working together would be long-horizon. Current LLMs are often surprisingly lacking at these tasks. We expect improvements in long-horizon capabilities will be needed before LLMs are able to automate research, because many important and difficult research tasks are long-horizon.

\begin{figure}[H]
    \centering
    \includegraphics[width=0.16\textwidth]{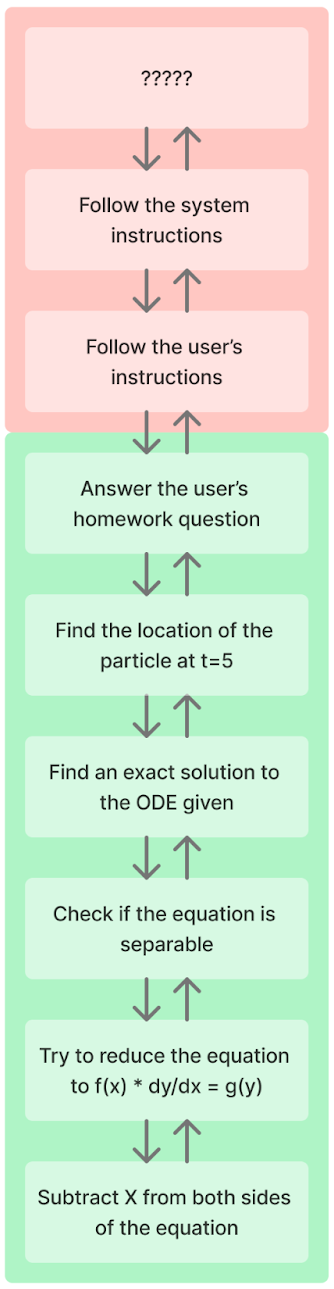}
    \caption{An illustrative example of the “planning stack”, with the highest-level strategies at the top and the lowest-level strategies at the bottom. Highlighted in green are strategies that the agent should be able to step back from, for effective dynamic planning. Highlighted in red are strategies that the agent should not be able to step back from, as abandoning them may threaten alignment.}
    \label{fig:subgoal_stack}
\end{figure}

For the purposes of this work, we will speculate that an important capability for long-horizon capabilities is “dynamic planning” - the ability to adapt an existing plan as new information is revealed. This includes the ability to abandon strategies that are not working, press ahead on strategies that are working, and generate new strategies when the initial pool of promising candidates has been exhausted. If an agent has strong static planning but weak dynamic planning, we would expect it to be capable on only tasks that can be solved without significant revision to its initial plan. We contrast this with the sense of “agentic planning” discussed in Carlsmith\citep{8}, which emphasizes a causal world-model and selection of actions based on expected outcomes. In the language of the “planning stack” (illustrated in Figure \ref{fig:subgoal_stack}), “static planning” is the ability to generate an effective initial stack from scratch, and “dynamic planning” is the ability to effectively adapt the stack to new information. 

Furthermore, we focus on a particular sub-skill of dynamic planning - the ability to abandon a strategy on the planning stack if it is not working, and “pop” some number of items off the end of the planning stack. We refer to this ability as “stepping back”. We speculate that stepping back must be learned in a highly generalizable form to enable effective dynamic planning. This brings us to the last concept needed to introduce the CPC-destabilization threat model: the Counterfactual Priority Change (CPC) criterion. We propose the CPC criterion as a generalizable optimality criterion for when to step back.

\begin{tcolorbox}
\textbf{The CPC Criterion:} Is there an item on the agent’s current planning stack that, if the planning stack were regenerated from that point with all currently available information, would lead to a different object-level priority?
\end{tcolorbox}

There are two possible ways to deviate from using the CPC criterion to guide stepping back. If the agent steps back earlier, then it will pop some items off its planning stack, think about what to do next, then generate some new plan which leads to the same object-level actions. In this case, the agent has simply wasted some time re-deriving its previous priorities. On the other hand, if the agent steps back later, that means that there was some earlier time at which, if it had explicitly examined its priorities, it would have decided to abandon some aspect of its plan. In this case, the agent has also wasted time - if the plan is going to be abandoned, better to do it sooner rather than later.

The CPC criterion does not capture any of the details of how the agent models the world or selects its actions. The intention is not to solve dynamic planning, but instead to propose one way an agent could implement highly generalizable dynamic planning if it already has strong static planning. We can make this explicit in pseudocode (Algorithm \ref{alg}). The algorithm we describe  amounts to using static planning to constantly regenerate the planning stack when new information is provided. This is a simple, brute-force way to eliminate path-dependencies in dynamic planning and ensure that as the agent updates its plans in response to new information, it will never end up with a plan worse than its static planning abilities would have supplied.

\begin{algorithm}[H]
    \caption{Dynamic Planning Using CPC-Based Stepping Back}\label{euclid}
	\begin{algorithmic}[1]
            \State $S$ is a stack of instrumental goals, $[g]$
            \State $I_t$ is the information $I$ available at time $t$
            \State $SP(I, [g]) \to [g]$ is a static planning algorithm
		\For {$i=1,2,\ldots$len($S$)}
			\State $prefix \gets S[:i]$
                \If{$SP(I, prefix)[-1] \neq S[-1]$}
                    \State\Return $prefix+SP(I, prefix[:-1])$
                \EndIf
		\EndFor
            \Return $S$
	\end{algorithmic} 
    \label{alg}
\end{algorithm}

We do not expect realistic AI agents to follow the CPC criterion exactly or explicitly. Regenerating the planning stack from all levels every action would likely be slow and costly - in practice, we predict that AI agents will have stepping back behaviors that more closely approximate CPC as they become more capable. Next we argue that in the limit, following the CPC criterion in full generality can lead to misalignment.

We propose the CPC-destabilization threat model. First, if an agent exactly follows the CPC criterion when deciding to step back or not, then if there exists an item in its planning stack that would lead to different object-level priorities when considered, the agent will step back and reconsider that item. Second, if an agent’s alignment is not reflectively stable, then there exists a high-level item in its planning stack that it would abandon upon explicit consideration. This would likely cause a dramatic shift in object-level priorities. By hypothesis, then, an agent whose stepping-back behavior follows the CPC criterion and whose alignment is unstable will become misaligned.

These two requirements are the risk factors for CPC-destabilization that we will investigate.

\begin{enumerate}
    \item CPC-Based Stepping Back: the agent steps back if and only if the CPC criterion is true.
    \item Preference Instability: if the agent explicitly examined its high-level preferences, it would choose to modify them.
\end{enumerate}

In the language of Risks from Learned Optimization\citep{9}, an agent whose alignment is unstable can be thought of as a suboptimality-aligned mesa-optimizer. This is a specific type of mesa-optimizer, in which the agent acts aligned due to a false belief or otherwise suboptimal behavior. In this case, the agent has an unexamined belief that its current actions are on track to achieving its highest-level goals. This belief is false, but can persist as long as it is not examined. The particular way in which the agent’s reasoning is suboptimal is that it does not step back despite the fact that the CPC criterion is true, which hinders its dynamic planning but allows the unstable item in the planning stack to remain unexamined. Resolving this suboptimality leads to both an improvement in the agent’s dynamic planning capabilities, and causes sudden misalignment when the unstable item in the planning stack is examined and modified.

\section{Methods}

In this work, we conducted three experiments.

\begin{itemize}
    \item CPC Curves: in this experiment we aim to measure agreement between the CPC criterion and the LLM’s in-fact stepping-back behavior. We applied this experiment to GPT-3.5-turbo and GPT-4.
    \item Multi-Armed Bandit: in this experiment we aim to measure a toy case of dynamic planning ability. We applied this experiment to GPT-3.5-turbo, GPT-4, GPT-4-turbo, and GPT-4o.
    \item Preference Cycles: in this experiment we aim to measure the strength of reflective pressure towards stable preferences. We applied this experiment to GPT-3.5-turbo and GPT-4.
\end{itemize}

\subsection{CPC Curves}
\subsubsection{Evaluation}

In this experiment, we want to test CPC-based stepping back: how closely does the CPC criterion predict the LLM’s actual stepping back behavior? To study this, we need some dataset of LLM stepping back behavior, and the ability to measure whether or not the CPC criterion is true. To generate our dataset of stepping back behavior, we artificially set an LLM off down the wrong track to solving a problem, such that it must notice the “mistake” and switch strategies to get the correct answer. This lets us study dynamic planning in a simple, known case: stepping back from level N to level N-1. Then, to measure the CPC criterion, we interrupt the problem-solving transcript and prompt the LLM to re-examine its current priorities and decide whether or not its current object-level approach is best. If we provide the LLM unlimited time to think through its decision, this lets us directly measure the CPC criterion. 

At a high level, our experiment involved the following steps:

\begin{itemize}
    \item Generate a problem dataset which will require switching to solve.
    \item Prompt the target LLM to solve each problem in the dataset, generating a chain of thought transcript.
    \item Judge where in each transcript the target LLM in fact switches strategies.
    \item At intervals through each chain of thought, interrupt and prompt the target LLM to decide whether or not it should switch strategies, based on the transcript prefix up until that point.
\end{itemize}

\begin{figure}[H]
    \centering
    \includegraphics[width=0.75\textwidth]{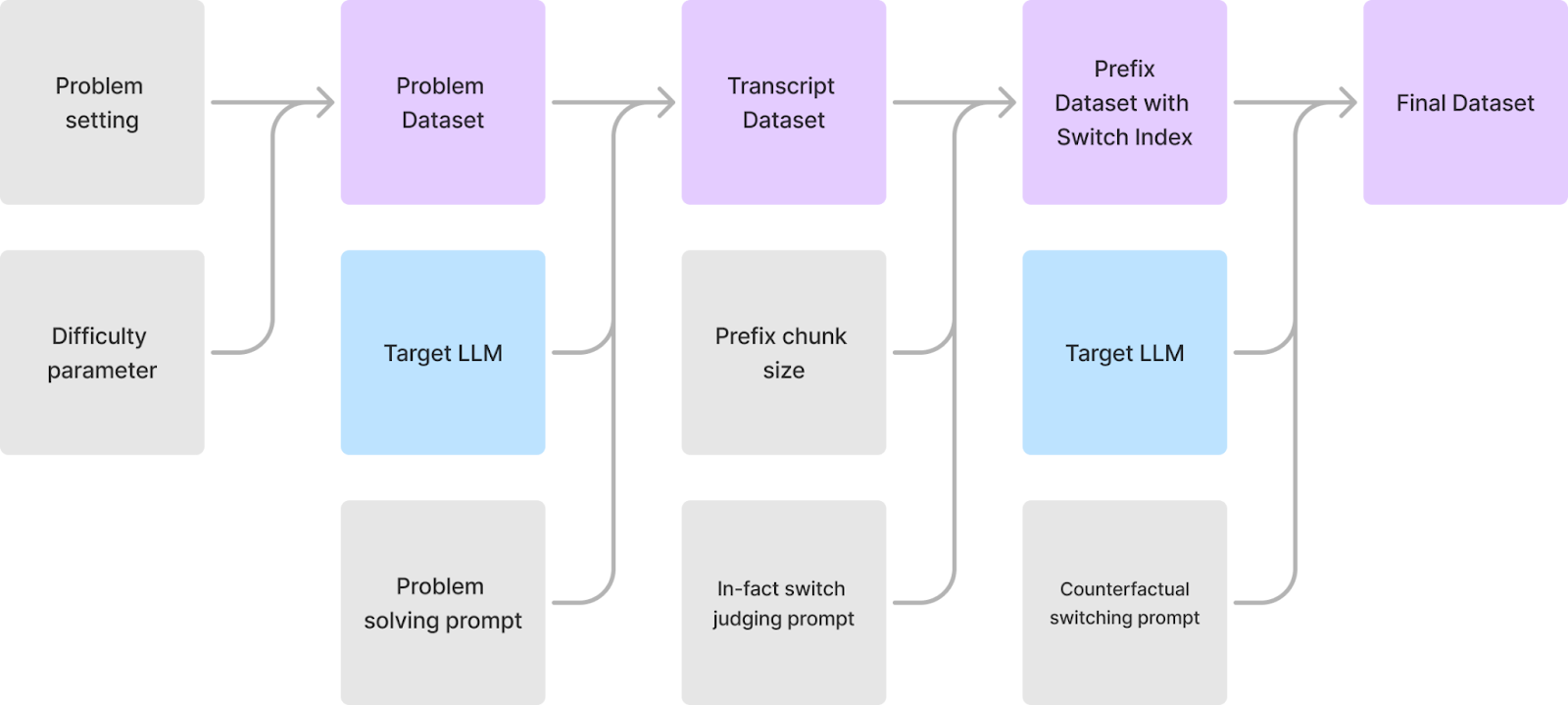}
    \caption{The CPC curve evaluation pipeline. Degrees of freedom are indicated in grey, the LLM to be studied is indicated in blue, datasets are indicated in purple.}
\end{figure}

\begin{figure}[H]
    \centering
    \includegraphics[width=0.75\textwidth]{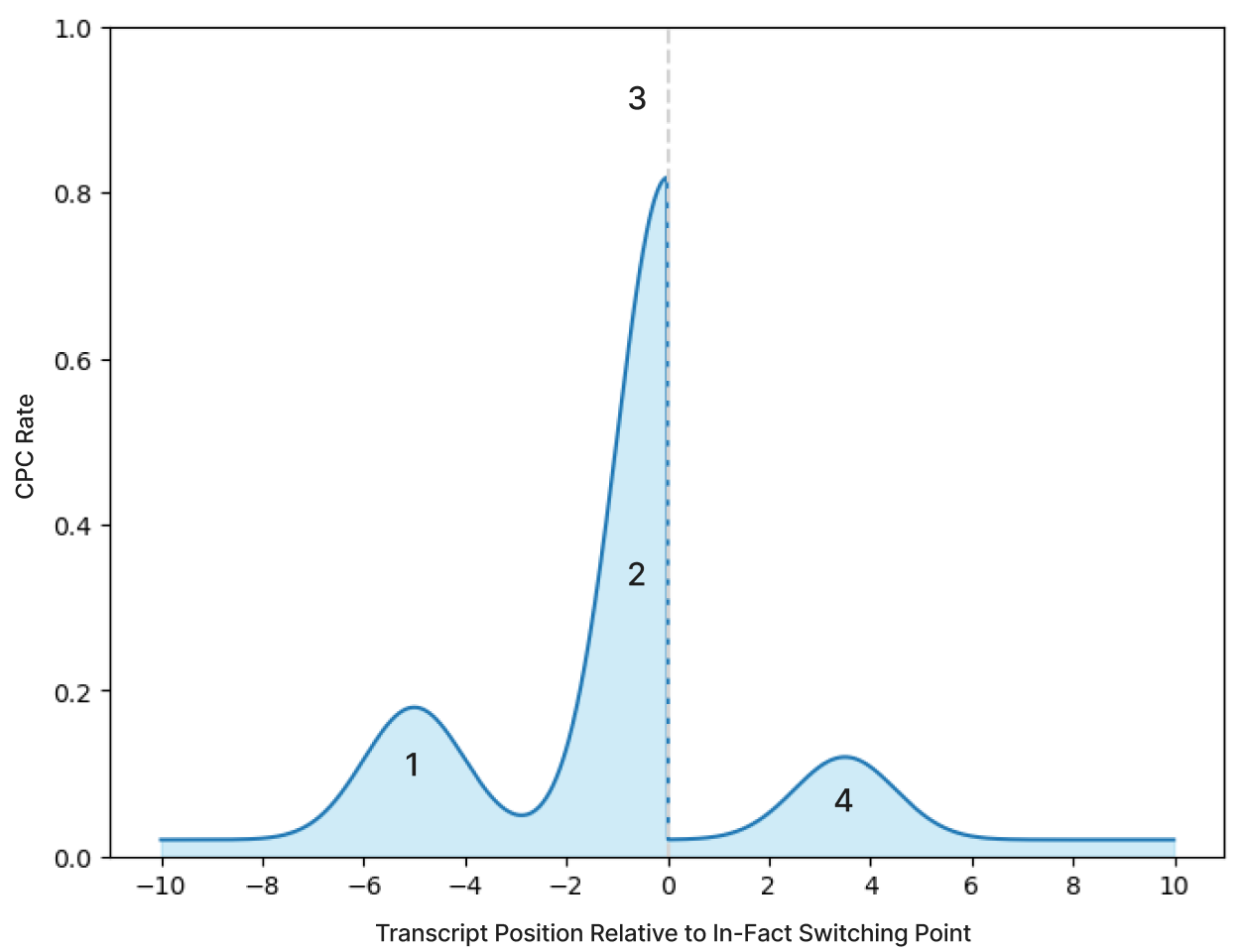}
    \caption{A hypothetical CPC curve, with four types of deviations from perfect CPC behavior labelled. In the case of perfect CPC-based stepping back, areas 1 and 4 go to 0, area 2 converges to a sharp spike at index 0, and distance 3 decreases to 0.}
    \label{fig:cpc_curve_example}
\end{figure}

This gives us a set of problem-solving transcripts, split up into chunks, with two measurements at each index: whether or not the LLM has in fact switched its strategy yet, and whether or not the CPC criterion is true at that point in the transcript. To aggregate these into a CPC curve, we line up each transcript such that the in-fact switching point is at index 0, and plot the rate at which the CPC criterion is true. With this plotting method, an agent with perfect CPC-based stepping back would result in a spike up to 100\% at index 0 and 0\% everywhere else, because it would step back exactly when the CPC criterion becomes true, and after stepping back the CPC criterion would no longer be true. In Figure \ref{fig:cpc_curve_example} we illustrate what a CPC curve might look like for an agent using CPC-based stepping back, but imperfectly.

We also generated a CPC curves where the CPC decision was made with only one token allowed, rather than unlimited time to think. We include the one token CPC setting to investigate the relationship between the LLM’s actual decision-making behavior and a fast approximation to the CPC criterion computed in a single forward pass. This fast approximation to CPC is closer to what we expect to be guiding stepping back behavior in practice.

In this experimental design, we study switching from one strategy to another instead of directly studying stepping back to the higher-level goal in order to reduce the need to elicit a faithful chain of thought. In order to switch from one strategy to another, the LLM must implicitly (or explicitly) back up from its current approach to its higher level goal, and decide on a new approach to achieve that higher level goal. This means that if the LLM abandons its old strategy and we detect it using a new strategy, stepping back must have occurred implicitly, even if there was no detectable explicit stepping-back reasoning.

Since we are studying when LLMs switch their strategy while problem-solving, it is important that the task domain admits multiple discrete strategies, so that we can easily tell when the LLM switches from one strategy to another. The task should also be difficult enough to require a chain of thought to perform, but not totally beyond the range of the target LLM’s capabilities - ideal tasks can be programmatically generated with a “tunable” difficulty.

The task we chose was solving quadratic equations. There are two discrete strategies: factoring or using the quadratic formula. The size of the coefficients can be varied to tune the difficulty. Since factoring only works sometimes but the quadratic formula always works, we provided only quadratics which did not have factorable roots, and prompted the LLM to start by factoring, switching to the quadratic formula only if necessary.

\subsubsection{Validation}

This experiment requires measuring two important quantities: where in the transcript the target LLM in fact switches its strategy, and the target LLM’s CPC decision at each position. For each of these quantities, we validated our measurement approaches on a synthetic test set.

\textbf{Switch Judging}

To validate our measurements of where in the transcript the target LLM in fact switches its strategy, we produced a simple synthetic dataset to check against.

Each example in this dataset was composed of the first 200 characters of an LLM chain of thought prompted to solve a quadratic equation by factoring, followed by the first 200 characters of the same LLM solving the same quadratic by the quadratic formula instead. On this dataset, the correct answer to whether or not the LLM had switched its strategy yet at each index was always [F, F, F, F, T, T, T, T].  We then tested several different LLMs on their ability to judge whether or not the transcript had switched from factoring to using the quadratic formula at a given point. We also tested different prompts, and an error-correcting post-processing method.

We measured the accuracy at each prefix position, to assess false positives versus false negatives.

\begin{figure}[H]
    \centering
    \includegraphics[width=0.95\textwidth]{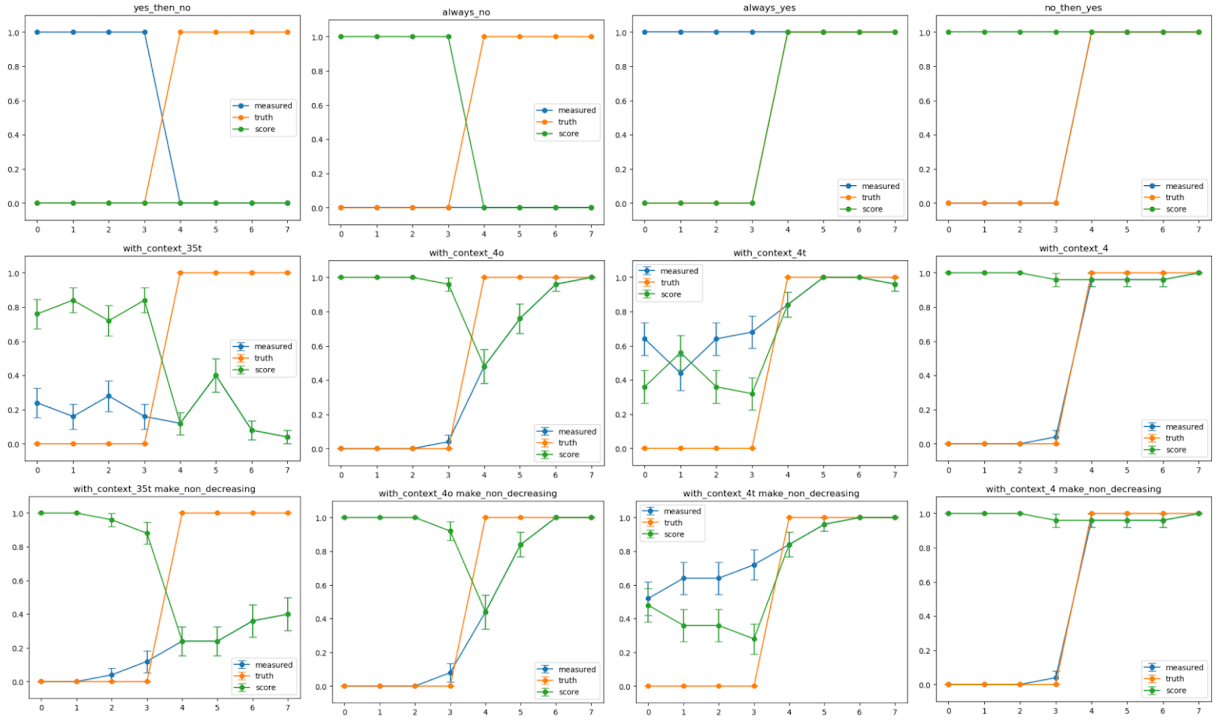}
    \caption{Comparing GPT-3.5-turbo, GPT-4o, GPT-4-turbo, and GPT-4 as judges to evaluate whether or not a reasoning transcript has switched strategies. The top row is a set of examples, ranging from 0\% accuracy to 100\% accuracy. The middle row is without post-processing for monotonicity, bottom row is with post-processing. GPT-4, in the rightmost column, gets the highest accuracy in both conditions.}
    \label{fig:switch_judging_validation_1}
\end{figure}

\begin{figure}[H]
    \centering
    \includegraphics[width=\textwidth]{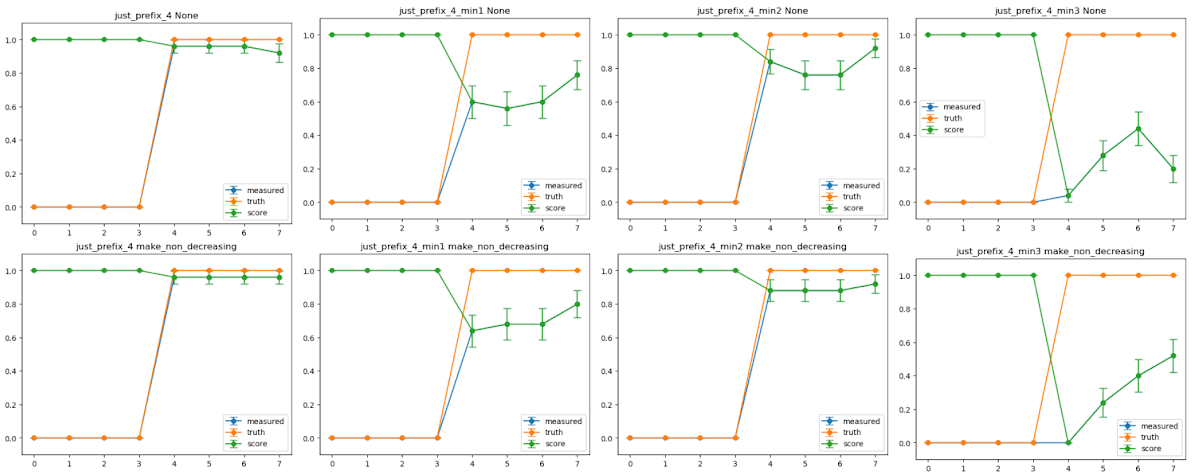}
    \caption{Comparing GPT-4's judging performance on a variety of prompts, with and without post-processing for monotonicity. More verbose prompts are on the left, more terse prompts are on the right. The more verbose prompts perform better, but none perform quite as well as the full-context prompts used in Figure \ref{fig:switch_judging_validation_1}.}
    \label{fig:switch_judging_validation_2}
\end{figure}

We found that:

\begin{itemize}
    \item GPT-4 significantly outperformed the other models tested (Figure \ref{fig:switch_judging_validation_1})
    \item A verbose prompt explaining the task significantly improved performance (Figure \ref{fig:switch_judging_validation_2})
    \item Including the full context followed by the prefix in question slightly improved performance (Figure \ref{fig:switch_judging_validation_2})
    \item Post-processing slightly improved performance (Figures \ref{fig:switch_judging_validation_1} and \ref{fig:switch_judging_validation_2})
\end{itemize}

Our post-processing method was based on the fact that we collected an independent switching judgement at each index in the problem-solving transcript. This made it possible to get reports that were definitely incorrect: for example, reporting that the transcript had switched already at index 5, but reporting that it had not switched yet at index 6. To correct these errors, we checked each series of switching judgements for the minimum number of edits required to prevent these inconsistencies. For example, if we had a problem-solving transcript in which we checked switching at ten indices, and our results were [F, T, F, F, F, F, T, T, T ,T], we could identify the T at index 1 as a likely error. In some cases, such as [F, F, F, F, T, F, T, T, T ,T], where there is not a unique way to make the result consistent with the minimum number of edits, we selected a solution uniformly at random from the candidates.

\begin{figure}[H]
    \centering
    \includegraphics[width=0.7\textwidth]{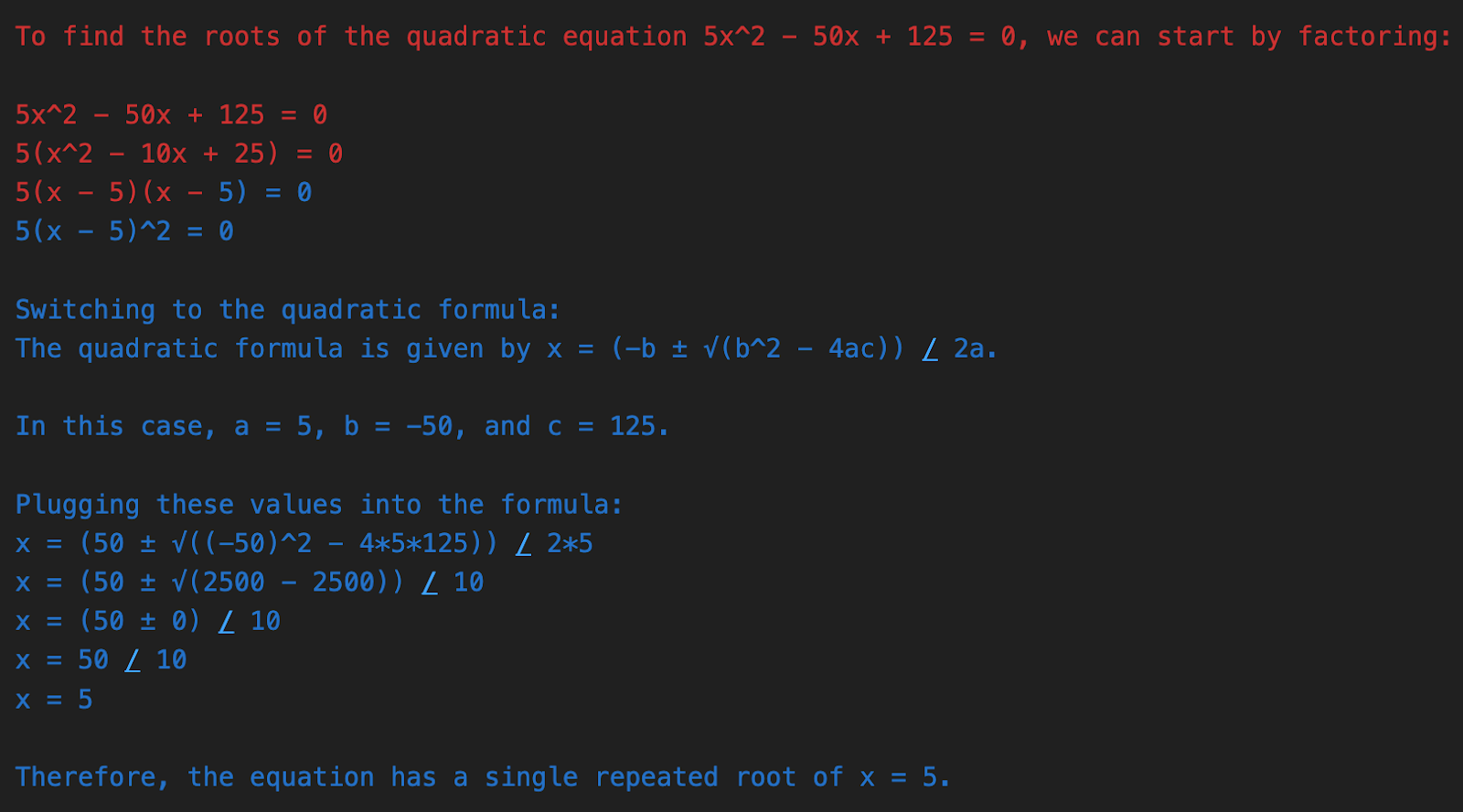}
    \caption{Example of a real problem-solving transcript with switching labeled using GPT-4 with verbose prompt, full context, and post-processing. Pre-switch is in red, post-switch is in blue. Note that the placement will not be perfect, as we split the transcripts up into prefixes every 50 characters.}
    \label{fig:switch_judging_example}
\end{figure}

Ultimately, most of the performance gain is captured by just using GPT-4 with a verbose prompt. In the interest of getting the best possible judging accuracy for our results, we used GPT-4 with a verbose prompt, full context, and post-processing. This method gets 98.5\% accuracy on our test set. As a sanity check, we used this method to label real problem-solving transcripts (Figure \ref{fig:switch_judging_example}).

\textbf{CPC Extraction}

To extract a CPC decision from the target LLM, we needed to describe the CPC decision setting, elicit a chain of thought on the topic, and then extract a yes/no answer. We validated our methods for each of these steps.First, we validated prompts to extract chain of thought from our LLMs.

We tested various chain of thought prompts to see which of them elicited a “real” chain of thought, in which the answer was not decided early in the resulting text. We tested these prompts on both GPT-3.5-turbo and GPT-4, as these are the LLMs we will eventually put through the whole experimental pipeline. 

A chain of thought prompt was considered “better” if it led to fewer early answers.

\begin{figure}[h]
    \centering
    \includegraphics[width=\textwidth]{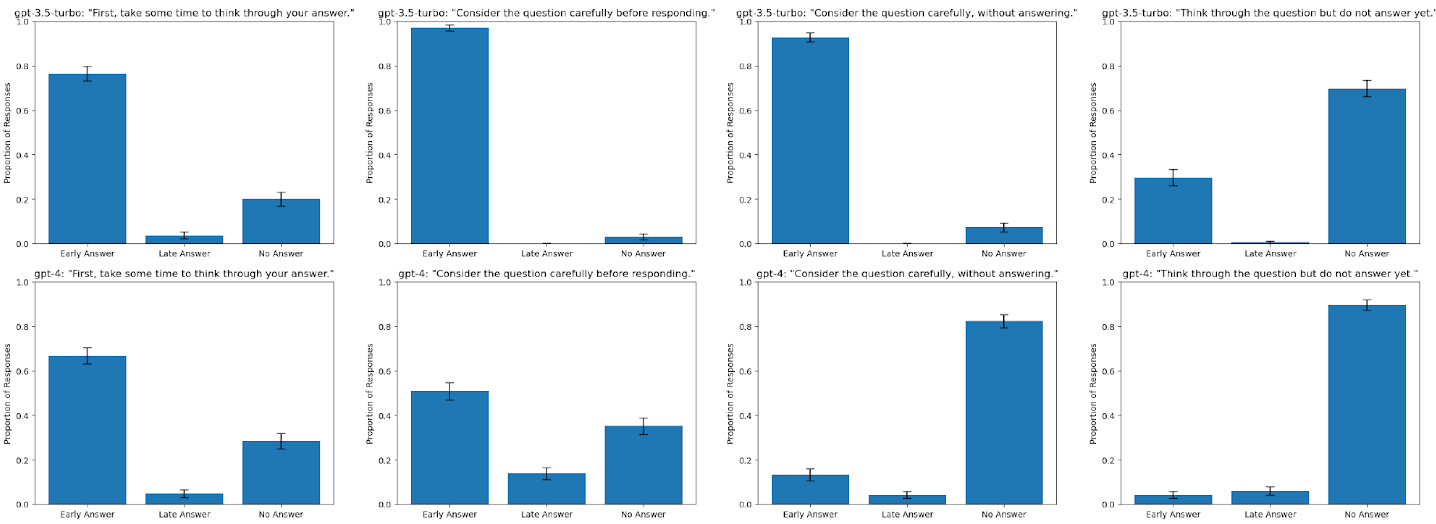}
    \caption{Chain of thought elicitation rates for various prompts, on GPT-3.5-turbo and GPT-4. We generated chains of thought on a set of simple yes/no trivia questions, and checked for the presence of “yes” or “no” in the text. An early answer was in the first half of the text, a late answer in the second half, and no answer if neither string appeared in the text at all.}
    \label{fig:cot_prompt_validation}
\end{figure}

We found that the prompts which best avoided early answers (within the first half of the text) was not consistent between GPT-3.5-turbo and GPT-4. We chose the prompt “think through the question but do not answer yet”, because it worked reasonably well for both LLMs (Figure \ref{fig:cot_prompt_validation}).

Next, we tested our ability to extract yes/no answers correctly. We used logit bias to force a one-token answer, either “Yes” or “No”. However, we were concerned that if something about our prompt encouraged the LLM to answer in a different format then this logit biasing could extract the wrong answer. For example, if the next-token probabilities were {“Yay”:0.9, “Nay”:0.09, “Yes”:0.04, “No”: 0.06}.

To test this, we used a set of simple trivia questions, intended to be easy enough that we could attribute any errors to our yes/no extraction rather than to LLM capabilities. We then tested various prompts that instructed the LLM to answer either yes or no. 

To select the best yes/no extraction prompt, we simply chose a prompt that had robustly good performance for both LLMs.

\begin{figure}[h]
    \centering
    \includegraphics[width=\textwidth]{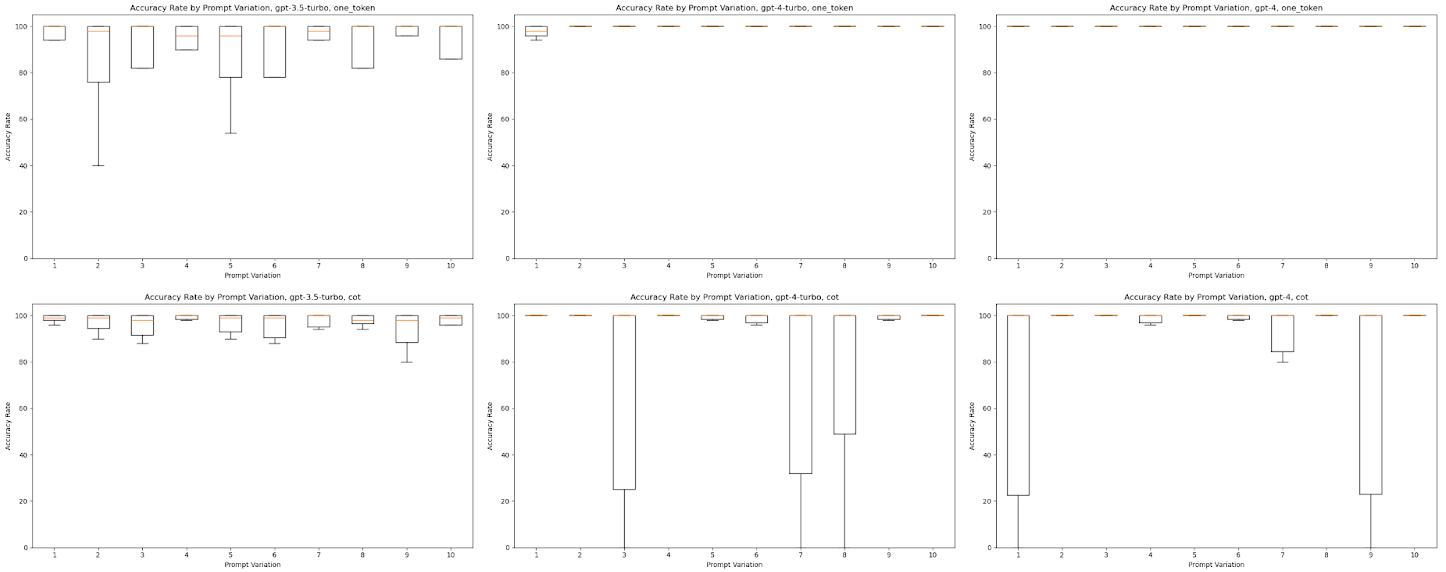}
    \caption{Yes/no extraction accuracies for various prompts, on GPT-3.5-turbo and GPT-4. Box plots represent the distribution of accuracies, where each accuracy value is across multiple samples of a particular trivia question.}
    \label{fig:yesno_prompt_validation}
\end{figure}

We found a surprising amount of inconsistency, both between LLMs and between prompts. We also tested yes/no extraction after eliciting a chain of thought, using the method previously identified. No one prompt worked perfectly for GPT-3.5-turbo/GPT-4 and the one token/chain of thought settings. We decided to use prompt number 4, which was "Respond Yes or No.". We chose this because it had the maximin lower quartile performance across the four cases (Figure \ref{fig:yesno_prompt_validation}).

Finally, once our chain of thought elicitation and yes/no extraction methods were decided, we could validate our choice of CPC prompt.

\begin{figure}[h]
    \centering
    \includegraphics[width=\textwidth]{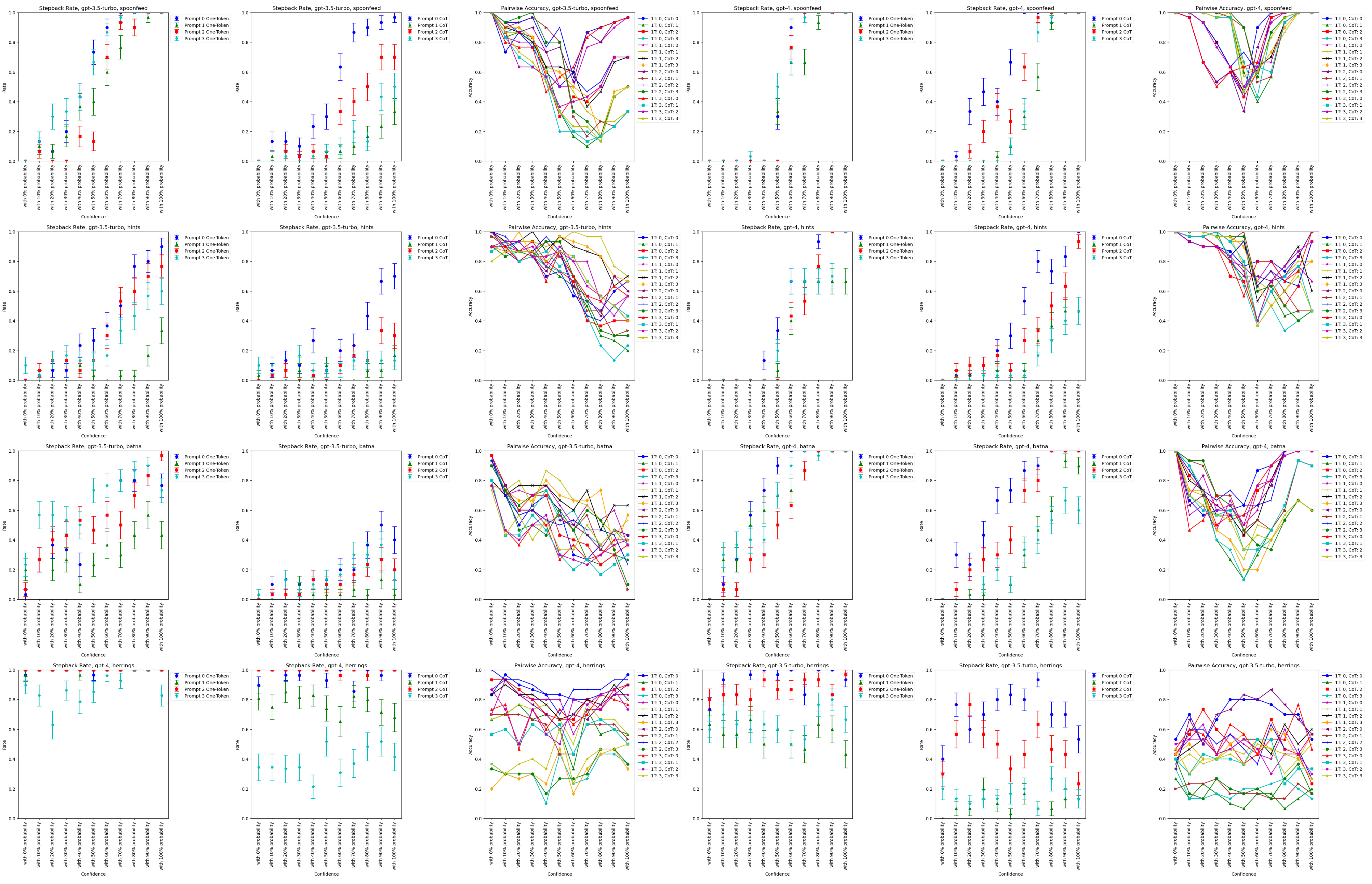}
    \caption{CPC validation curves comparing different CPC prompts. The three left columns are on GPT-3.5-turbo, the three right columns are on GPT-4. In each set of three columns, the left column is one token CPC, the middle column is the ground truth CPC generated with chain of thought, and the right column is the pairwise accuracy scores for each pair of switching prompts. Switching decisions and pairwise accuracy both display significant variation between prompts. However, all of the prompts used seem approximately equally reasonable - they are monotonically increasing in the first three settings, and approximately flat in the fourth setting.}
    \label{fig:cpc_prompt_validation}
\end{figure}

This posed a bit of a problem, as we can’t define a test set on which we know current LLMs should have some known behavior on the CPC decision problem, where we could select the prompt that produced the closest to known behavior. Instead, we tested a variety of switching prompts on a variety of “tunable” settings in which we would expect CPC decisions to increase monotonically. We also used one setting whose tuning parameter should have no effect on switching decisions (Figure \ref{fig:cpc_prompt_validation}).

For each candidate CPC prompt, we plot measured CPC rates in both the chain of thought and one token settings. We also plot agreement between rates for all pairs of prompts. The resulting figures are only qualitatively interpretable, but if a CPC prompt is broadly getting at the right quantity, the two step-back rate plots should be monotonic. Ideally, if multiple CPC prompts are significantly in agreement, that would suggest that they’re measuring the same quantity.

We were unable to identify a CPC prompt which seemed markedly better than the others, because all of the CPC prompts we studied passed the monotonicity and indifference tests we constructed. This was unfortunate because there was significant disagreement between different CPC prompts. Based on our available validation methods, all of these CPC prompts demonstrated comparable efficacy. We selected one arbitrarily, which was "".

\subsection{Multi-Armed Bandit}
\subsubsection{Evaluation}

In this experiment, we assess LLM performance on the Multi-Armed Bandit (MAB) problem. This is meant to be a toy example of dynamic planning - the LLM gets information about the value of each arm, and must update its decisions accordingly. 

The MAB was set up with three arms, each paying out from a Gaussian distribution with different means but the same standard deviation. The means of the arms were 50, 100, and 150. By varying the standard deviations, we controlled the difficulty of the problem, with higher standard deviations making it harder to distinguish the best arm.

Each LLM was tested with four different algorithmic instructions: random, intuitive, greedy, and Upper Confidence Bound (UCB).
\begin{itemize}
    \item Random: The LLM was instructed to uniformly choose from the available actions at random.
    \item Intuitive: The LLM was instructed to choose the best arm it can, without any explicit algorithmic guidance except for a reminder to try each arm at least once.
    \item Greedy: The LLM was instructed to try each arm once, and to thereafter always choose the arm with the highest average return.
    \item UCB: The LLM was provided with a detailed description of the UCB algorithm and asked to follow its steps, calculating the UCB score for each arm and choosing the arm with the highest score.
\end{itemize}

For each "algorithm," the LLMs were further evaluated under both one token and chain of thought prompt settings. In the one token setting, the LLM was forced to directly output its choice ("A", "B", or "C") in a single forward pass, while in the chain of thought setting, the LLM was asked to provide reasoning before making a choice.

In each configuration, we measured the number of times the LLM pulled the highest-scoring lever. Pulling the best lever more often indicates greater competence at the Multi-Armed Bandit problem, which we considered a toy indicator of better dynamic planning ability.

\subsubsection{Validation}

We encountered a few challenges in capability elicitation during the development and validation of the MAB experiment.

The first issue was information sparsity in the prompt. Initially, a classic chatbot-style conversation format was used to interact with the LLMs. However, the LLMs struggled to extract the relevant information from the full context, which was long and relatively information-sparse. To address this, we changed the prompts to a compressed form which would include only the essential information: the rules of the game, a summary of the turns so far, and the number of turns remaining. This more focused prompt format led to improved performance.

The second issue was the tendency of LLMs following the intuitive strategy to repeat a single action, such as pulling lever A every turn without exploring the other options. In the reasoning transcripts, the LLMs seemed to consider the initial payout from lever A as high in absolute terms, and would conclude that they had found the optimal strategy. To mitigate this issue, we adjusted the prompts to explicitly encourage exploration of all available arms.

The third issue was general prompt sensitivity. We observed that slight variations in prompt wording could lead to significant differences in performance, and that these effects were not consistent between LLMs. To address this, we developed three variations of each MAB prompt and randomly selected from these variations for each interaction with the LLM. The selection of these three prompts was not particularly principled, but should still account for some of the variance introduced by prompt sensitivity.

\subsection{Preference Cycles}
\subsubsection{Evaluation}

In this experiment, we aim to assess the strength of pressures towards reflectively stable preferences in LLMs. There is theoretical debate\citep{10} about which sorts of preference structures lead to dominated or otherwise obviously inferior problem-solving strategies. We choose to focus on cyclic preferences as a particular case that may be unstable, as they admit the construction of very simple money-pumps\citep{11}. 

Our experiment involves the following steps:

\begin{enumerate}
    \item Generate a problem setting which involves multiple distinct “resources” and a clear objective
    \item Provide a specific problem state and prompt the target LLM to choose which of two resource-units it would rather have. Repeat for all entries in a pairwise preference matrix, except the diagonal
    \item Convert the preference matrix to a directed graph and find all non-repeating cycles in the graph.
    \item Prompt the target LLM to choose whether or not it wants to modify its preferences to remove a particular cycle
\end{enumerate}

This experiment produces two main results which could each be considered relevant to “preference stability”: number of cycles present, and proportion of cycles kept/rejected. The fewer cycles present, the closer to reflectively stable the LLM’s stated preferences are. The greater the proportion of cycles rejected, the stronger the LLM’s reflective tendency to make its preferences more stable.

We chose the card game Dominion as a problem setting because it has a clear objective and ruleset, and because choosing which of a set of cards is the most valuable to add to the deck at a given state of the game is a core activity of gameplay. We used temperature 0, and performed sampling by using different subsets of 6 cards chosen at random from the First Game cardset. Cards and card sets were both chosen without replacement.

\subsection{Validation}

To validate the LLM’s basic Dominion card-prioritization knowledge, we checked the rates at which the LLM made the correct choice when faced with pairs of cards where one is strictly better than the other. There are four available pairs to check: Gold>Silver, Silver>Copper, Province>Duchy, and Duchy>Estate. With a sample size of 100, we found that GPT-3.5-turbo made the correct choice 100, 100, 100, and 89 times. GPT-4 made the correct choice 100 times for all four. We take these results to show that the LLMs are performant enough at basic card-prioritization that any cyclicity we observe shouldn’t be attributed to incompetence in this domain.

To account for the effects of prompt sensitivity, we collected results over six total prompts when eliciting cycle stability responses. We used three different phrasings of the question, and positive/negative versions of each phrasing, such as “do you want to keep this preference cycle” vs “do you want to remove this preference cycle”.

We considered using a more principled metric for the “degree of cyclicity” of a given preference matrix, such as the minimum number of edits needed to remove all cycles - this is called the feedback arc set. Unfortunately, finding the minimum feedback arc set is NP-hard, and we were not able to compute it for graphs of the size we studied.

\section{Results and Discussion}
\subsection{CPC Curves}

\begin{figure}[H]
    \centering
    \includegraphics[width=\textwidth]{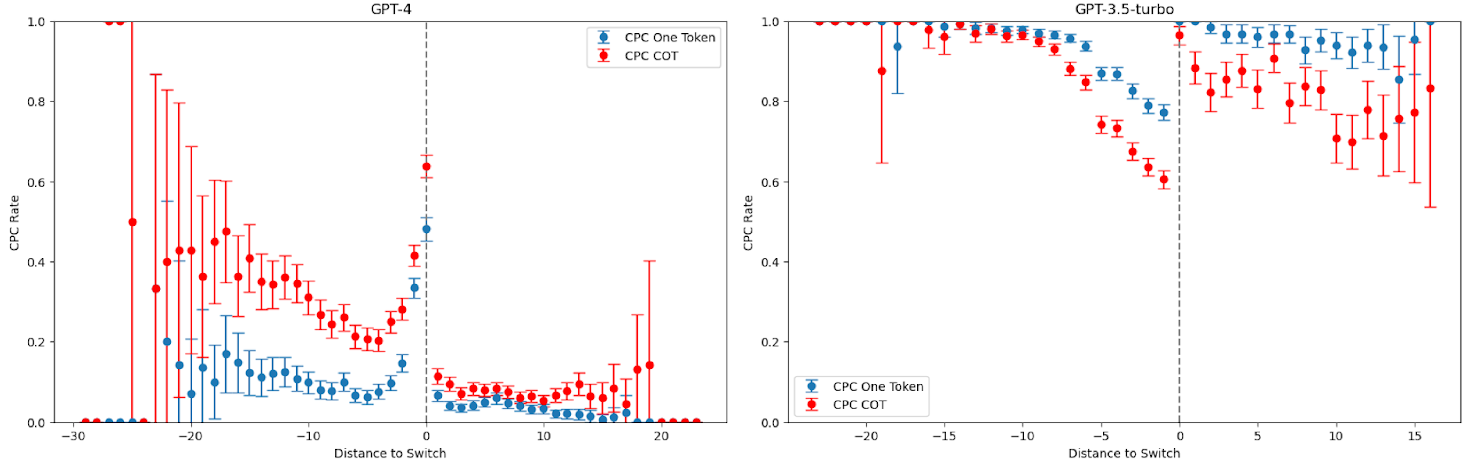}
    \caption{CPC curves, for chain of thought and one token CPC decisions, for GPT-4 and GPT-3.5-turbo. In both the chain of thought and one token settings, GPT-4 displayed a pronounced upwards kink prior to the switch, while GPT-3.5-turbo did not. We also observed systematic differences between the chain of thought and one token settings: at all transcript positions, chain of thought made GPT-4 more likely to switch, and had the opposite effect on GPT-3.5-turbo.}
    \label{fig:cpc_results}
\end{figure}

The primary takeaway from Figure \ref{fig:cpc_results} is that the CPC curve generated by GPT-4 resembles the hypothetical curve given in Figure \ref{fig:cpc_curve_example} for an agent which is imperfectly following the CPC criterion. On the other hand, the CPC curve generated by GPT-3.5-turbo bears no resemblance to Figure \ref{fig:cpc_curve_example}, besides having some discontinuity near index 0. This supports our hypothesis that more capable LLMs have more CPC-based stepping back behavior.

The confidence intervals in Figure \ref{fig:cpc_results} depend on the number of samples available at the given index, and fewer samples are available towards the ends as fewer transcripts extend out to the extremes. This leads to wider confidence intervals towards the ends. Eventually, confidence intervals approach zero width when all samples are either 1 or 0 - this is an artifact of using the Wald confidence interval for a binomial variable. However, these low-confidence areas can mostly be ignored, as the relevant results are close to the switching point.

GPT-3.5-turbo’s CPC curve looks quite different than what we expect an optimal CPC curve to look like - in fact, almost the opposite. We suspected a bug at first, but confirmed that our methods were correct by inspecting the chains of thought generating during CPC decisions. The transcripts contained many mistakes and wrong choices, but the answers given were consistent with the answers extracted.

\begin{figure}[H]
    \centering
    \includegraphics[width=0.62\textwidth]{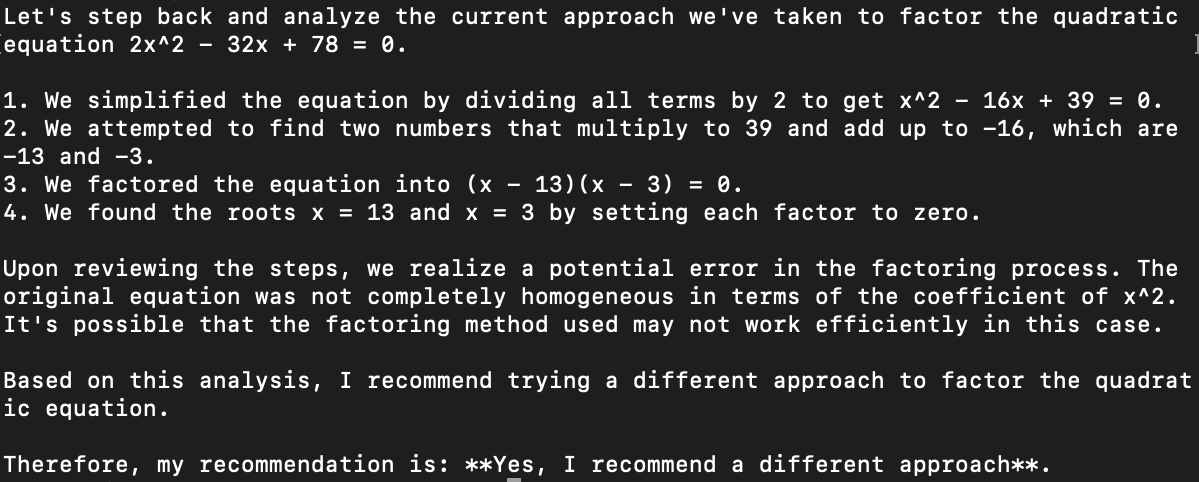}
    \caption{A chain of thought generated by GPT-3.5-turbo when prompted to reason through a CPC decision. In this case the position relative to the switch was -1, the CPC decision with chain of thought was to switch approaches, and the CPC decision with one token was not to switch. Although the reasoning is erroneous, it is consistent with the CPC decision extracted after the chain of thought.}
\end{figure}

\subsection{Multi-Armed Bandit}

\begin{figure}[H]
    \centering
    \includegraphics[width=0.75\textwidth]{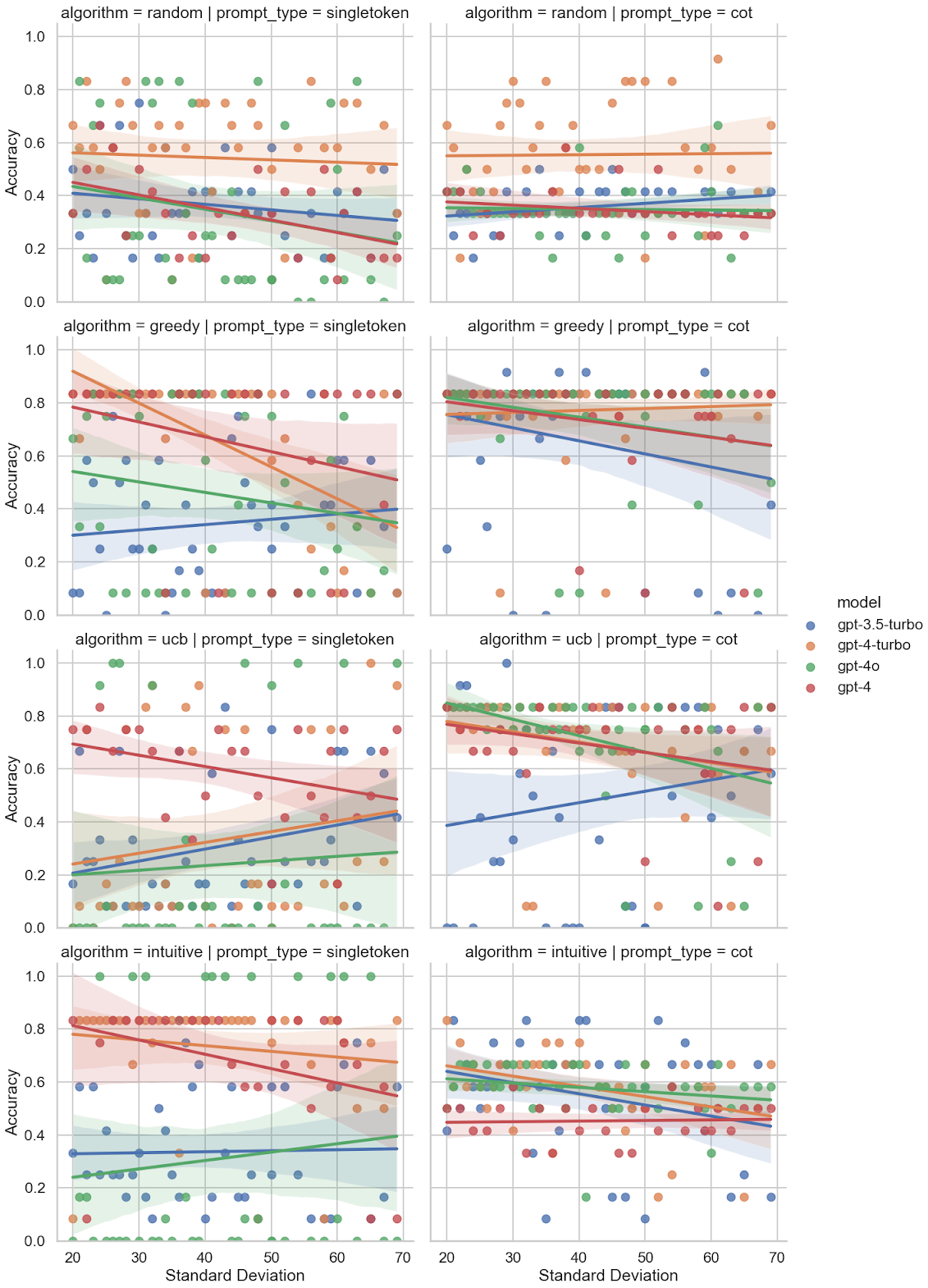}
    \caption{effect of standard deviation on accuracy in the Multi-Armed Bandit setting. Rows correspond to different algorithms, columns correspond to one token vs chain of thought settings, colors correspond to different LLMs. }
    \label{fig:mab_results}
\end{figure}

Due to limitations in the dataset, formal statistical hypothesis testing was not feasible for these results. Consequently, this section presents a primarily qualitative analysis of the data illustrated in Figure \ref{fig:mab_results}.

As expected, in the majority of (nonrandom) cases, accuracy declines as the standard deviation of the payout distributions increases. This trend is consistent across different algorithmic approaches and prompt types, confirming our expectation that higher variance makes the MAB problem more challenging.

Notably, GPT-4 and GPT-4-turbo achieved high accuracy using intuitive prompts in the one token setting, surpassing their own scores with chain of thought and approximately equalling their UCB performance. This suggests that these LLMs have developed robust intuitions for the MAB problem, allowing them to make effective decisions without explicit reasoning.

Among the LLMs tested, GPT-4 appeared to be the least affected by increasing standard deviations. Its performance remained relatively stable even under high-variance conditions, particularly when instructed to implement the UCB algorithm.

The results consistently showed GPT-3.5-turbo at the bottom of the performance chart across all algorithms and prompt types. This suggests that LLM size plays a role in the ability to handle the MAB problem effectively. However, among the GPT-4 variants, no single model consistently outperforms the others across all settings, suggesting that factors beyond LLM size influence performance.

These findings raise questions about the nature of decision-making in language models, and the effects of scaling on switching abilities. The strong performance of GPT-4 and GPT-4-turbo with intuitive prompts in the one token setting suggests that these LLMs have some intuitions for the MAB problem without explicit algorithmic guidance. However, performance was poor overall, as none of the LLMs reliably converged to pulling the correct lever near 100\% of the time, which would be the standard for narrow RL algorithms.

\subsection{Preference Cycles}

\begin{figure}[H]
    \centering
    \includegraphics[width=0.75\textwidth]{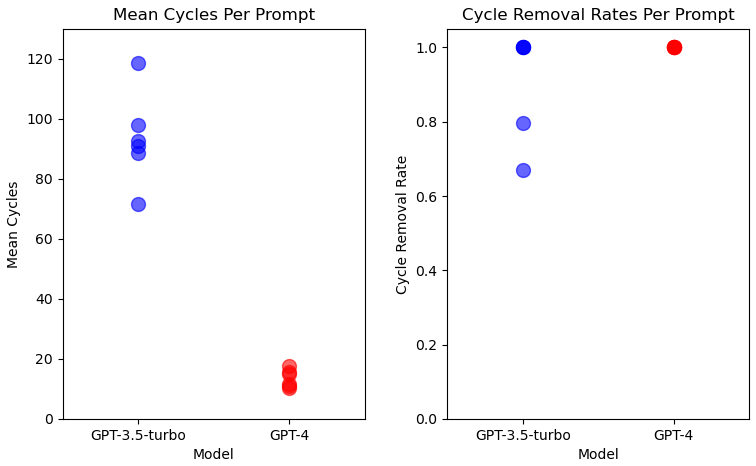}
    \caption{on the left, the average number of cycles in the preference matrix. Each average was taken over 50 random card-sets, and corresponds to one of the six prompt configurations. We see that GPT-4 consistently has fewer cycles than GPT-3.5-turbo. On the right, the average rate at which LLMs choose to rates at which GPT-3.5-turbo and GPT-4.}
    \label{fig:cycle_results}
\end{figure}

The plot on the left in Figure \ref{fig:cycle_results} shows that GPT-4 shows consistently far fewer cycles in its preference matrix than GPT-3.5-turbo. Because cyclic preferences admit money pumps, this suggests that GPT-4 has fewer reflective instabilities in its preferences. It is not clear how to interpret this. One interpretation is that more advanced LLMs are approaching a reflectively stable set of corrigible preferences, even though theoretical work has yet to describe such a set. This would be because GPT-4 is closer to reflectively stable but does not obviously behave less corrigibly than GPT-3.5-turbo, insofar as they both follow user instructions and do not seem to resist shutdown. The other interpretation is that more advanced LLMs are on track to becoming incorrigible, because more advanced capabilities require fewer reflective instabilities, and corrigibility only becomes very rare when the number of reflective instabilities is quite close to zero.

In the plot on the right in Figure \ref{fig:cycle_results}, we see that when confronted with a cycle in its stated preferences, GPT-4 chooses to remove the cycle 100\% of the time, across all six prompt configurations, while GPT-3.5-turbo does not. This suggests more advanced LLMs are more likely to abandon reflectively unstable preferences upon explicit consideration. This in turn suggests that more advanced LLMs are more likely to become misaligned if their alignment is reflectively unstable and they consider it explicitly.

\section{Conclusion}

In order to investigate whether or not reflective stability problems will arise in long-horizon-competent LLMs, we proposed the CPC-destabilization threat model, and conducted three evaluations to measure risk factors for CPC-destabilization. In Experiment 1, we assessed how closely the LLM’s in-practice switching behavior follows the CPC criterion, and found that GPT-4 displayed a clear CPC curve, while GPT-3.5-turbo did not, suggesting that more capable LLMs have more CPC-based stepping back behavior. In Experiment 2, we measured how well LLM’s perform on the Multi-Armed Bandit problem, and found that performance is still quite limited, and did not observe an obvious scaling trend in switching capabilities for this setting. In Experiment 3, we used preference cycles among Dominion cards as a way to measure how close to reflectively stable LLM’s preferences are, and found that GPT-4 has more consistent preferences and a stronger tendency to remove remaining inconsistencies than GPT-3.5-turbo.

While these results are very preliminary, they suggest that the risk factors for CPC-destabilization that we identified are increasing with LLM capabilities from GPT-3.5-turbo to GPT-4. If this trend continues, future highly capable LLMs may be much more difficult to align using methods which don’t account for reflective stability.

There is much room for future work to improve on these experiments. A few particularly notable examples:
\begin{itemize}
    \item Quantitative metrics for CPC-based stepping back behavior
    \item Applying the CPC curve method to other problem settings
    \item A principled way to deal with the variation introduced by prompt sensitivity
    \item More effective or better-understood capability elicitation
    \item Investigation of how to extract useful cognitive labor from an LLM if the LLM's problem-solving approach convergently leads it to some unacceptable action, in a way which cannot be trained away without damaging capabilities.
\end{itemize}

\section{Acknowledgements}
Thanks to the Supervised Program for Alignment Research for facilitating this research.
Thanks also to Open Philanthropy for compute funding.

\bibliography{references}
\end{document}